\documentclass[conference]{IEEEtran}
\IEEEoverridecommandlockouts

\usepackage{cite}
\usepackage{amsmath}
\usepackage{graphicx}
\usepackage{multirow}
\usepackage{enumerate}
\usepackage{amsmath,amssymb,amsfonts}
\usepackage{algorithmic}
\usepackage{subfigure}
\usepackage{textcomp}
\usepackage{xcolor}
\usepackage{color}
\usepackage{booktabs}
\usepackage{enumerate}
\usepackage{threeparttable}
\usepackage{enumitem}
\usepackage{graphicx}
\usepackage{mwe}
\usepackage{indentfirst}
\usepackage{soul}
\usepackage{threeparttable}

\def\BibTeX{{\rm B\kern-.05em{\sc i\kern-.025em b}\kern-.08em
    T\kern-.1667em\lower.7ex\hbox{E}\kern-.125emX}}
\begin{document}

\title{A Novel Meta Learning Framework for Feature Selection using Data Synthesis and Fuzzy Similarity\\
% \thanks{Identify applicable funding agency here. If none, delete this.}
}

\author{\IEEEauthorblockN{Zixiao Shen, Xin Chen, Jonathan M. Garibaldi}
	\IEEEauthorblockA{\textit{Intelligent Modelling and Analysis Group, School of Computer Science} \\
	    \textit{Lab for Uncertainty in Data and Decision Making (LUCID)} \\
		\textit{University of Nottingham}, Nottingham, NG8 1BB, United Kingdom \\
		\{Zixiao.Shen, Xin.Chen, Jon.Garibaldi\}@nottingham.ac.uk}
}

% use for special paper notices
\IEEEspecialpapernotice{Accepted by IEEE World Congress on Computational Intelligence (WCCI), 19-24th July, 2020, Glasgow, UK}

\maketitle

\begin{abstract}
This paper presents a novel meta learning framework for feature selection (FS) based on fuzzy similarity. The  proposed method aims to recommend the best FS method from four candidate FS methods for any given dataset. This is achieved by firstly constructing a large training data repository using data synthesis. Six meta features that represent the characteristics of the training dataset are then extracted. The best FS method for each of the training datasets is used as the meta label. Both the meta features and the corresponding meta labels are subsequently used to train a classification model using a fuzzy similarity measure based framework. Finally the trained model is used to recommend the most suitable FS method for a given unseen dataset. This proposed method was evaluated based on eight public datasets of real-world applications. It successfully recommended the best method for five datasets and the second best method for one dataset, which outperformed any of the four individual FS methods. Besides, the proposed method is computationally efficient for algorithm selection, leading to negligible additional time for the feature selection process. Thus, the paper contributes a novel method for effectively recommending which feature selection method to use for any new given dataset.
\end{abstract}

\section{Introduction}
Due to rapid development and wide application of information technology, an increasing number of datasets with high dimensions and complexity are continually being generated. Much research work focuses on knowledge and pattern extraction using different machine learning models \cite{han2011data}. In the data mining area, feature selection (FS) acts as a pre-processing strategy to reduce the dimensionality and redundancy of data. It has played an essential role for preventing the problems of overfitting, reduction in the computational cost of applying machine learning algorithms to datasets and for enabling comprehensive decision making \cite{shen2018performance}.

In the literature, there are a number of different FS methods. Based on the dependency with a learning algorithm, FS methods can be generally grouped into three types, i.e. filter, wrapper and embedded methods \cite{chandrashekar2014survey}. Filter methods are independent of any learning algorithms, and have high computational efficiency compared with the wrapper and embedded methods. Hence, various filter FS methods are implemented and compared in this research.

One of the main issues for applying different FS methods is that the performance of the various FS methods varies in a manner which is data dependent. That is, it is not possible to state categorically which is the optimal FS method, in the sense of providing the best performance for all kinds of data \cite{hua2009performance}. This situation poses an interesting and challenging problem as to how to select the best FS method to use for any given unseen dataset \cite{parmezan2017metalearning}.

One approach of solving this problem is through ensemble or combination methods \cite{shen2019novel}. By combining the diversity kinds of FS algorithms, ensemble methods result in a better performance by taking advantages of different methods. On the other hand, the combination process may suffer from a high computational cost. In some cases, ensemble methods do not necessarily achieve a better performance than any of the individual methods.

Another approach is using meta-learning method to choose the best algorithm for a given dataset \cite{brazdil2008metalearning}. Meta learning learns the selection of the most appropriate FS method for a given dataset in a meta level. It is normally trained to learn the relationship between the characteristics of training datasets and their corresponding best FS methods \cite{parmezan2017metalearning}. This is quite valuable and of great importance in the decision making area.

In order to deal with the impreciseness and uncertainty in such a decision making context, fuzzy sets and fuzzy methods have been developed to model many practical problems \cite{zadeh1965fuzzy}. A fuzzy similarity based framework has been utilized to solve the classification problems in a flexible and explainable way \cite{luukka2006similarity}. A comprehensive evaluation of the fuzzy similarity based framework has been reported in our previous research \cite{shen2018performance}.

In the current paper, a novel meta-learning method is proposed to achieve automatic selection of the best FS method for a given dataset using a fuzzy similarity based framework. The rest of the paper is organised as follows. Section~II presents a more detailed literature review of feature selection methods; Section~III presents the methodology used; Section~IV presents experiments in which the new methodology is applied to a range of real world datasets, and the paper closes with Conclusions.

\begin{figure*}[tb]
	\centering
	\includegraphics[width=1.8\columnwidth]{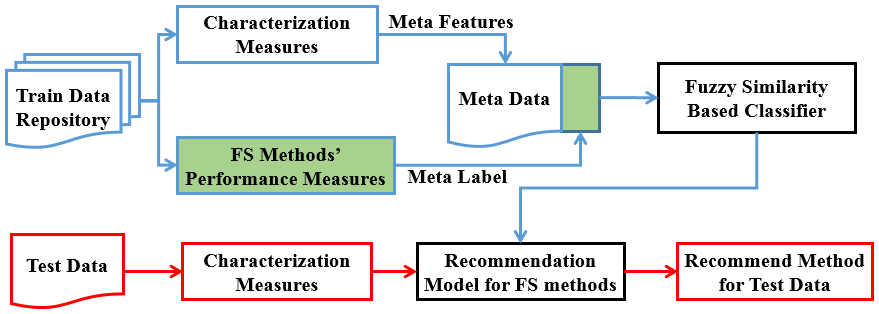}
    % 	\decoRule
	\caption[Framework of Fuzzy Sets Generation using Bootstrap Sampling]{Overall framework of the proposed architecture. Blue lines and red lines show the data flows for training and testing processes respectively.}
	\label{overall_framework}
\end{figure*}

\section{Literature Review}
Meta learning, in our scenario, is defined as a process of learning the meta-knowledge to improve model learning using machine learning and data mining methods \cite{brazdil2008metalearning}. Nowadays, it is becoming a hot topic to improve the stability and generalization of the learned models. There are two main aspects of research in meta-learning methods. One is algorithm selection, which aims to choose the best algorithm based on learning the relationship between the characteristics of the datasets and the performances of different algorithms \cite{kalousis2001feature}. The other one focuses on parameter selection, which aims to determine the optimal parameters of a sophisticated FS method \cite{reif2012meta}.

Meta learning methods for algorithm selection have been applied to solve different problems, such as classification, regression, optimization, time series prediction, etc. \cite{lemke2015metalearning, kuck2016meta, lemke2010meta}. Several researchers have also investigated the application of meta learning method for feature selection \cite{parmezan2017metalearning, filchenkov2015datasets}. In general, there are two mainly approaches to algorithm selection, namely feature engineering and neural architecture search. Recent research on meta learning tend to focus on the neural architecture search, which is powerful, generic and versatile to different models. On the other hand, it can also lead to the high computational cost when exploring the suitable configurations. Comparatively, feature engineering is an efficient process which extracts the meta features from the raw dataset \cite{hartmann2019meta}. In this research, a feature engineering based meta learning process has been explored here. Several issues are discussed and investigated as below.

\paragraph{Construction of A Data Repository for Training}
The central concept of meta learning is to learn the knowledge from a data repository. Hence, the selection and construction of the data repository becomes quite essential. In other studies, a large number of real-world datasets have been used to construct a meta database. Parmezan et al. \cite{parmezan2017metalearning} have used 150 real datasets for the meta-learning process. However, it is quite a time consuming process to collect various kinds of datasets from real-world applications and this is often restricted by ethical issues. More importantly, these real datasets may not cover a wide range of characteristics that are similar to the given unseen dataset under consideration. In this paper, we propose to use synthesized datasets to construct a large data repository that covers a variety of characteristics for meta learning.

\paragraph{Selection of Meta Features}
The meta features refer to the features which have certain relationships with the algorithm performance. The selection of meta features is also dependent on the treated problem. In \cite{reif2014automatic}, the meta features are generally classified into five groups: simple, statistic, information theoretic based, model based and land-marking. A number of widely used meta features are implemented in our study.

\paragraph{Choice of A Recommendation Method} 
To construct the meta model, different decision making methods have been used, such as decision trees \cite{parmezan2017metalearning}, support vector machine, kNN, etc. Using fuzzy methods to perform the recommendation has been rarely reported before. So on the basis of our previous work \cite{shen2018performance}, we will implement a fuzzy similarity based framework to achieve the decision making in this paper.

The main contribution of the current paper is the implementation of a meta learning method for feature selection using fuzzy similarity measure. More importantly, instead of using real datasets for meta-learning (e.g. \cite{parmezan2017metalearning}), we propose to use data synthesis to generate a large number of datasets that cover a wide range of characteristics, leading to a more generalized meta-learning solution. Our method has been subsequently evaluated on eight public datasets of real-world applications. It achieves a superior performance in recommending the best FS method for each dataset.

\section{Methodology}
As illustrated in Fig. \ref{overall_framework}, the proposed method mainly consists of five steps: (1) Generation of a data repository for training; (2) Meta features extraction; (3) FS methods' performance measures; (4) Meta data construction; (5) Recommendation modeling using fuzzy similarity measure. Blue lines and red lines in Fig. \ref{overall_framework} show the data flows for training and testing processes respectively. In the training phase, the meta features and meta labels have been generated using the synthetic datasets. After implementing different FS methods on the synthetic datasets, the meta label is obtained to represent the FS method with the best performance. A set of meta features are extracted to represent the characteristics of the synthetic datasets. A fuzzy similarity-based classifier is also introduced during the decision making process. In the testing phase, the same meta features are extracted from the test dataset. By applying the recommendation model in the training phase, the recommended optimal method for the test dataset is obtained. The detailed information are described in the following subsections.

\subsection{Generation of a Data Repository for Training}\label{configure_data}
In this section, we aim to construct a data repository that covers a variety of characteristics using data synthesis. Comparing with other synthetic datasets, the Madelon dataset \cite{guyon2008feature} holds the advantages of high flexibility and variability. It consists of relevant, redundant, repeated and useless features. The dataset presents a wide range of values for the number of features and samples. In addition, the values can also be distorted by adding noise, flipping labels, and shifting and rescaling processes \cite{bolon2013review}. By implementing the methodology as first proposed in the NIPS 2003 feature selection challenge \cite{guyon2003design}, different kinds of madelon datasets are generated for our proposed method by varying 11 different parameters, as listed in Table~\ref{madelon_parameters}. The value range for each of these parameters is also listed in Table~\ref{madelon_parameters}. We set the value range for each parameter as large as possible to cover different scenarios. Note that in this paper, we only consider a binary classification problem, but it can be extended to a multiple class situation. Details of our experimental settings are described in Section~\ref{training_repository}.

\begin{table}[tb]\scriptsize
\caption{Parameters for data synthesis using Madelon dataset}
\centering
\begin{tabular}{c|c|c}
\hline
\textbf{Alias}  & \textbf{Meaning}   & \textbf{Value Range}  \\ \hline
P1  & Number of Classes  &   2  \\ \hline
P2  & \begin{tabular}[c]{@{}c@{}}Number of Useful Features\\ \textit{(initially drawn to explain the concept)}\end{tabular}  
& [4, 5,..., 20] \\ \hline
P3  & \begin{tabular}[c]{@{}c@{}}Number of Redundant Features\\ \textit{(linearly dependent upon the useful features)}\end{tabular}  
& [0, 1,..., 20] \\ \hline
P4  & \begin{tabular}[c]{@{}c@{}} Number of Repeated Features\\ \textit{(repeating P2 and P3 at random)}\end{tabular}    
& [0, 1,..., 20] \\ \hline
P5  & \begin{tabular}[c]{@{}c@{}} Number of Useless Features\\ \textit{(Drawn at random regardless of class label)}\end{tabular}
& [0, 1,..., 20] \\ \hline
P6  & Number of Samples per Cluster & [10, 11,..., 70] \\ \hline
P7  & Number of Cluster per Class  & [2, 3,..., 7] \\ \hline
P8  & Random Seed       & [1, 2,..., 1000]   \\ \hline
P9  & Factor multiplying the hypercube dimension  & [2, 3,..., 10]  \\ \hline
P10 & Fraction of y labels to be randomly exchanged & [0.01, 0.02, ..., 0.1] \\ \hline
P11 & Flag to enable or disable random permutations & [0, 1] \\ \hline
\end{tabular}
\label{madelon_parameters}
\end{table}

\subsection{Meta Feature Extraction}
To learn meta features from the synthetic dataset, we extract a set of meta features from $M$ different datasets $D_i$, $i=1,..., M$, each with the number of $S_i$ data samples ($E_1, E_2,..., E_{S_i}$) and $N_i$ features ($F_1, F_2,..., F_{N_i}$). The label information is represented using class $C$ ($c_1, c_2,..., c_{S_i}$) for different data samples. An overall description of the data structure is shown in Table~\ref{description_data}.

\begin{table}[tb]
\centering
\caption{Description of the structure of a generated synthetic dataset $D_i$ for meta feature extraction}
\begin{tabular}{c|c|c|c|c|c}
\hline
\multirow{2}{*}{\textbf{Samples}} & \multicolumn{4}{c|}{\textbf{Features}}                                  & \multirow{2}{*}{\textbf{Class}} \\ \cline{2-5}
                                  & \textbf{$F_1$} & \textbf{$F_2$} & ... & \textbf{$F_{N_i}$} &                                 \\ \hline
\textbf{$E_1$}  & $v_{11}$   & $v_{12}$  & ... & $v_{1N_i}$ & $c_1$ \\ \hline
\textbf{$E_2$}  & $v_{21}$   & $v_{22}$  & ... & $v_{2N_i}$ & $c_2$ \\ \hline
\textbf{...}    & ...        & ...       & ... & ...        & ...   \\ \hline
\textbf{$E_{S_i}$}  & $v_{S_i1}$  & $v_{S_i2}$   & ... & $v_{S_iN_i}$   & $c_{S_i}$  \\ \hline
\end{tabular}
\label{description_data}
\end{table}

Subsequently the six meta feature extraction methods, which are derived from each of the $D_i$ datasets, are described as below~\cite{parmezana2016supplementary}.

\begin{enumerate}% [fullwidth, itemindent=2em, label=\alph*)]
    \item \textbf{\textit{Number of Samples (NS)}}:  \\
    It represents the number of samples for each dataset.
    
    \item \textbf{\textit{Number of Features (NF)}}: \\
    It represents the number of features for each dataset.
    
    \item \textbf{\textit{Average Asymmetry of Features (AAF)}}: \\
    It measures the average value of the Pearson's asymmetry coefficient. The formulation is used to quantitatively summarize the skewness of a distribution which is shown in (\ref{aaf}).
    
    \begin{equation}\label{aaf}
        AAF(D_i) = \frac{3}{N_i} \sum^{N_i}_{j=1} \frac{ Mean(F_j) - Median(F_j)}{Std(F_j)} 
    \end{equation}
    
    where, $Mean(F_j)$, $Median(F_j)$ and $Std(F_j)$ indicate the average, median and standard deviation values of feature $F_j$ respectively. $j$ is the index of the features.
    
    \item \textbf{\textit{Average Correlation between Features (ACF)}}: \\
    It measures the average value of Pearson's correlation coefficient between different features.
    
    \begin{equation}
        ACF(D_i) = \frac{2}{N_i(N_i-1)} \sum^{N_i-1}_{j=1} \sum^{N_i}_{k=j+1} Pearson(F_j, F_k)
    \end{equation}
    
    where $Pearson(F_j, F_k)$ indicates the Pearson's correlation between feature $F_j$ and feature $F_k$.
    
    \item \textbf{\textit{Average Coefficient of Variation of Features (ACVF)}} \\
    It measures the average coefficient of variation by the ratio of the standard deviation and the mean of the feature values.
    \begin{equation}
        ACVF(D_i) = \frac{1}{N_i} \sum^{N_i}_{j=1} \frac{Std(F_j)}{Mean(F_j)}
    \end{equation}
    
    \item \textbf{\textit{Average Entropy of Features (AEF)}} \\
    This measures the average amount of the information that each feature provides for the prediction of the class.
    
    \begin{equation}
        AEF(D_i) = \frac{1}{N_i} \sum^{S_i}_{k=1}Entropy(F_j)
    \end{equation}
    
    where $Entropy(F_j)$ measures the distribution's entropy of feature $F_j$.
\end{enumerate}

\subsection{Performance Measures of FS Methods}\label{FS_Label}
In this step, we generate a label for each of the synthetic datasets, so that the derived meta features and their associated labels can be used for model learning to recommend the best FS method for a given unseen dataset. In this case, the label is the best FS method for a given training dataset. To determine the best FS method for each synthesized dataset, we firstly define a performance measurement metric based on classification accuracy, which is described as below.

FS methods are ultimately used to improve the classification accuracy using reduced number of features by removing redundant features. The FS methods we implemented in our study are able to rank the features from the most significant to the least significant. Then a classification model learning method is used to calculate the classification accuracies by gradually eliminating the least important features one at a time. The detailed procedures are listed as below.

\begin{enumerate}[itemindent=0em, label=\alph*)]
    \item Divide the data ($D_i$) into training sets and testing sets in a 10-fold cross validation manner;
    \item Implement the candidate FS method to rank the features using the training set;
    \item Based on the gradually reduced number of features, model the classifier using the training set and make the prediction on the test set (logistic regression is used as the classifier in our experiments);
    \item For each reduced number of features, calculate mean classification accuracy across different folds.
\end{enumerate}

\begin{figure}[tb]
	\centering
	\includegraphics[width=\columnwidth]{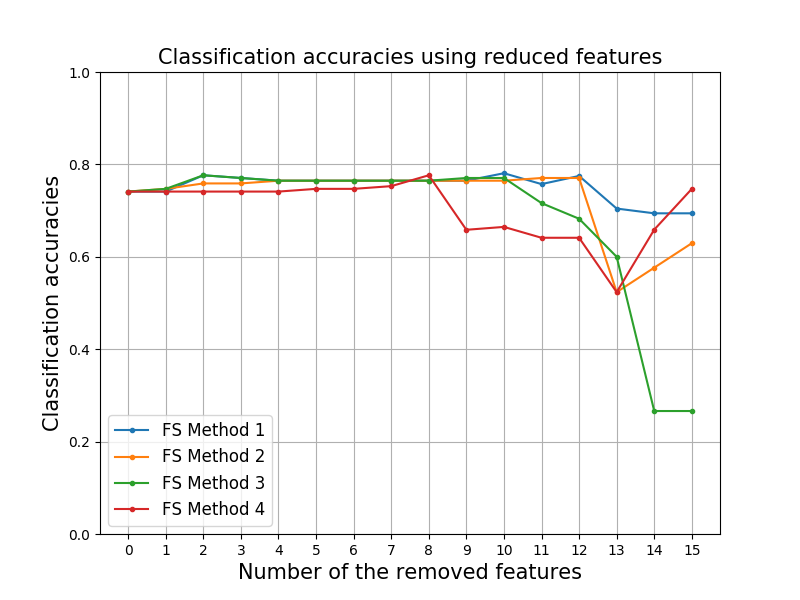}
	%\decoRule
	\caption[Framework of Fuzzy Sets Generation using Bootstrap Sampling]{Demonstration of classification accuracies using reduced features}
	\label{demonstration_acc}
\end{figure}

Fig. \ref{demonstration_acc} shows an example of plots of the classification accuracies obtained by increasing the number of removed features using different FS methods. Specific FS methods used in our study are described in Section \ref{FS_methods}.

As shown in Fig. \ref{demonstration_acc} that the classification accuracies vary significantly when features are gradually removed. A single measurement value is desirable to determine the best FS method. Either the mean or maximum value of the classification accuracy can be used, but these are not reliable due to the noisy nature of the curve. Here, we propose a new measure that is a weighted sum (WS) of the classification accuracies based on different numbers of removed features, as expressed by:
\begin{equation}\label{weighted_acc}
    WS = \sum Acc. * \% RemovedFeatures
\end{equation}
where $\% RemovedFeatures$ represents the proportion of the removed features and $Acc.$ means the corresponding classification accuracy using the retained features. The classification accuracies are calculated by removing features from the least significant to the most significant, hence the retained features become more and more important. As expressed in (\ref{weighted_acc}), we assign a higher weight (larger $\%RemovedFeatures$) to the classification accuracy that obtained from using more important features. If the features are ranked correctly by a FS method, a higher weighted sum should be achieved. This method is more robust to noise than the mean and maximum values.

\subsection{Meta Data Construction}\label{metadata_construction}
Based on the WS measure, we then select the FS method with the highest WS value as the meta label for the particular input dataset. Subsequently, the meta data is constructed by combining the six different meta features $MF_p$, ($1 \leq p \leq 6$) and the corresponding meta label for each dataset $D_i$. 
Based on the meta dataset, a decision making model is then trained to recommend the optimal FS method for a given dataset.

\subsection{Recommendation using Fuzzy Similarity Measure}
\begin{figure}[tb]
	\centering
	\includegraphics[width=\columnwidth]{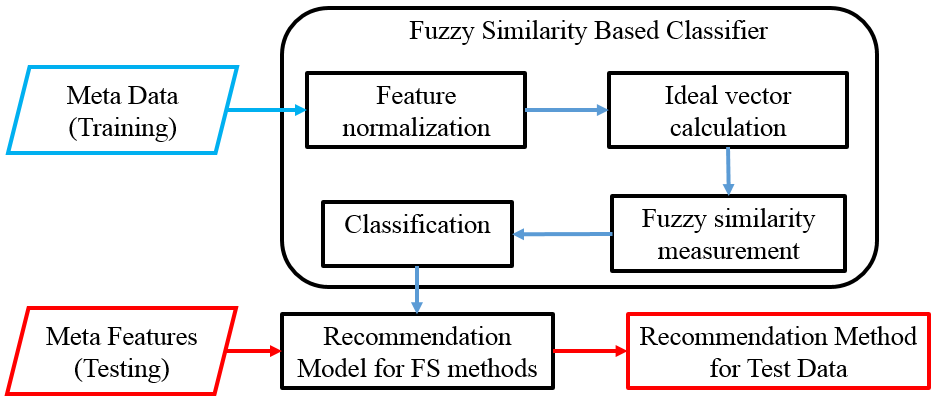}
	%\decoRule
	\caption[Framework of Fuzzy Sets Generation using Bootstrap Sampling]{Framework of fuzzy similarity based classifier. Blue and red lines show data flows for training and testing processes respectively \cite{shen2018performance}.}
	\label{similarity_framework}
\end{figure}

Based on our previous work \cite{shen2018performance}, a fuzzy similarity measure based framework is implemented to train a classification model using the generated meta dataset. The overall structure of the classification framework is illustrated in Fig.~\ref{similarity_framework}.

\begin{figure*}[tb]
	\centering
	\begin{minipage}[t]{0.3\linewidth}
		\centering
		\includegraphics[width=1.2\textwidth]{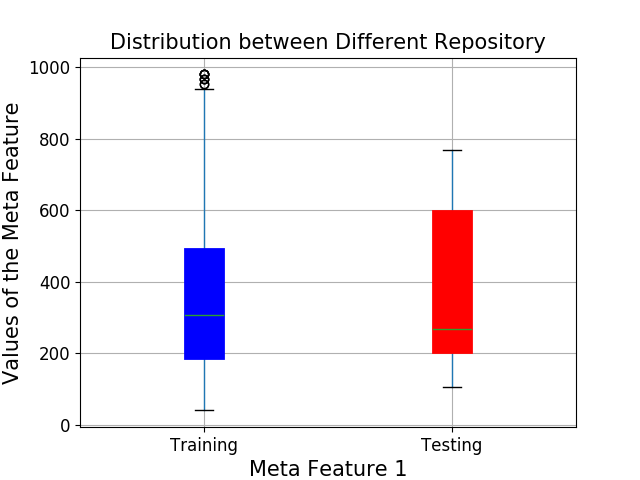}
		\parbox{1cm}{\small \hspace{3.5cm}(a){NS}}
	\end{minipage}
	\hspace{3ex}   %%两个minipage之间相隔3个字符的距离
	\begin{minipage}[t]{0.3\linewidth}
		\centering
		\includegraphics[width=1.2\textwidth]{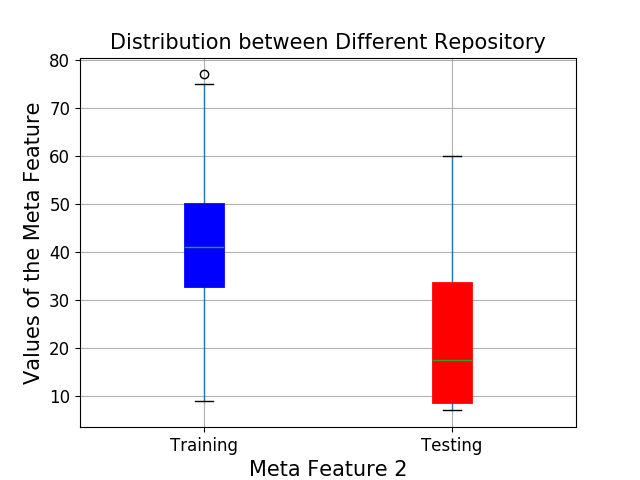}
		\parbox{1cm}{\small \hspace{3.5cm}(b)NF}
	\end{minipage}
	\hspace{3ex}   %%两个minipage之间相隔3个字符的距离
	\begin{minipage}[t]{0.3\linewidth}
		\centering
		\includegraphics[width=1.2\textwidth]{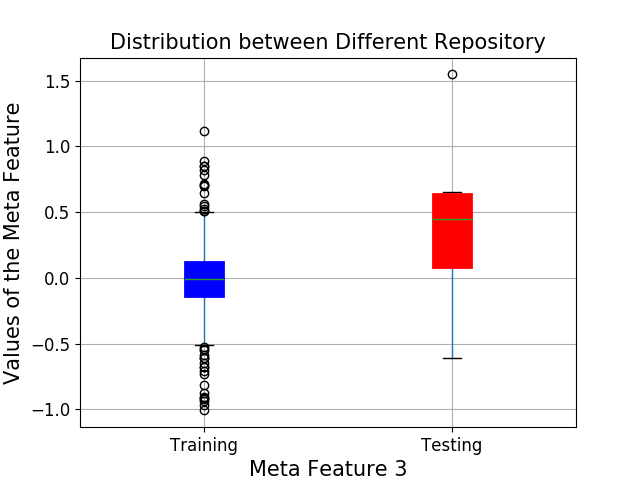}
		\parbox{1cm}{\small \hspace{3.5cm}(c)AAF}
	\end{minipage}
    \hspace{3ex}
	\begin{minipage}[t]{0.3\linewidth}
		\centering
		\includegraphics[width=1.2\textwidth]{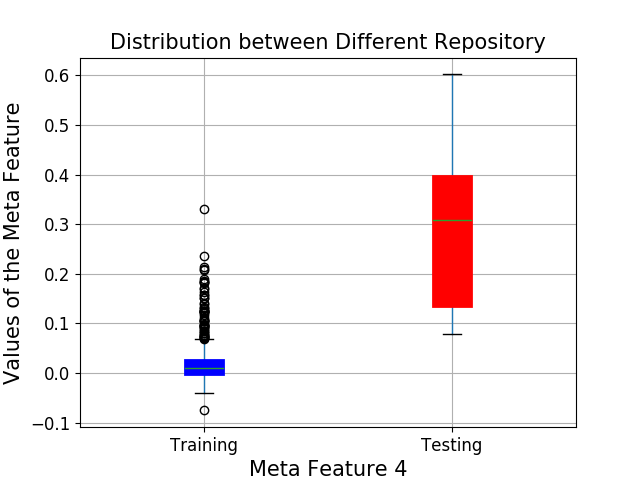}
		\parbox{1cm}{\small \hspace{3.5cm}(d){ACF}}
	\end{minipage}
	\hspace{3ex}   %%两个minipage之间相隔3个字符的距离
	\begin{minipage}[t]{0.3\linewidth}
		\centering
		\includegraphics[width=1.2\textwidth]{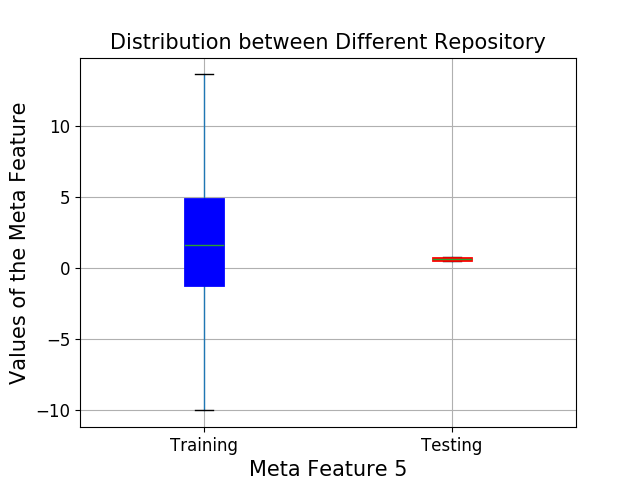}
		\parbox{1cm}{\small \hspace{3.5cm}(e)ACVF}
	\end{minipage}
	\hspace{3ex}   %%两个minipage之间相隔3个字符的距离
	\begin{minipage}[t]{0.3\linewidth}
		\centering
		\includegraphics[width=1.2\textwidth]{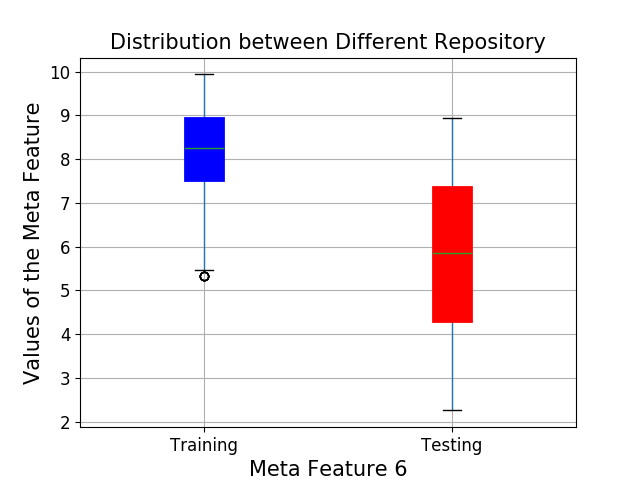}
		\parbox{1cm}{\small \hspace{3.5cm}(f)AEF}
	\end{minipage}
	\caption{Comparison of the distribution between the training and testing repository}
	\label{compare_distribution}
\end{figure*}

The model training process aims to classify a total number of $M$ examples $D_i$ ($1 \leq i \leq M$) into $L$ different classes $FS_l$, ($1 \leq l \leq L$) by their feature vector $\vec{x_q}$. $q$ is the index of the data samples in each class. $Z_l$ is the number of data samples for the $l$th class. Based on the comprehensive performance evaluation reported in \cite{shen2018performance}, the following fuzzy similarity measure based framework is implemented.

\begin{enumerate}[itemindent=1em, label=Step \arabic*:]
	\item For the training set, standardize each feature using the Z-score normalization process \cite{grus2019data}.
	
	\item Based on the standardized values from Step 1, calculate the ideal vector $\vec{v}_l$ for the $l^{th}$ class using geometric mean.
	\begin{equation}
    \vec{v}_l(p) = \sqrt[Z_l]{\prod_{q=1}^{Z_l}\vec{x_q}(p)}, \  1 \leq p \leq 6
    \end{equation}
    where $p$ represents the index of meta features.
    
	\item The same standardization process from Step 1 is  applied to the meta features extracted from the test dataset. Subsequently, feature vector $\vec{y_r}$ of the meta features is obtained, while $r$ indicates the index of the data samples in the test set.
	
	\item Based on the maximal fuzzy similarity measures proposed in \cite{luukka2001classifier}, a similarity measurement in the form of generalized \L ukasiewicz algebra is used. Geometric mean is used to combine the similarity measures from different features which is expressed in (\ref{eqGeo}).
	\begin{equation}\label{eqGeo}
    S \langle \vec{y_r}, \vec{v}_l \rangle = \sqrt[6]{\prod_{p=1}^{6} \sqrt{1 - |\vec{y_r}(p)^2- \vec{v}_l(p)^2|}}
    \end{equation}
    where $S \langle \vec{y_r}, \vec{v}_l \rangle$ represents the fuzzy similarity value between the feature vector of the testing set and the ideal vectors obtained from the training set.
    
    \item Classify the test dataset into the class with the corresponding ideal vector which produces the highest fuzzy similarity value.
    
\end{enumerate}

Through the steps above, the recommended FS method for a given test dataset is obtained.

\begin{figure*}[tb]
	\centering
	\begin{minipage}[t]{0.3\linewidth}
		\centering
		\includegraphics[width=1.2\textwidth]{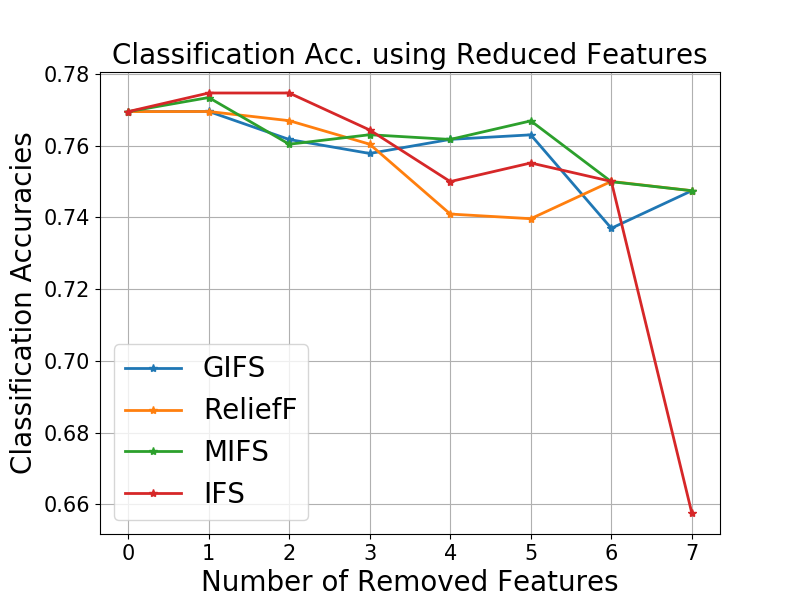}
		\parbox{1cm}{\small \hspace{4.5cm}(a)PIMA}
	\end{minipage}
	\hspace{3ex}   %%两个minipage之间相隔3个字符的距离
	\begin{minipage}[t]{0.3\linewidth}
		\centering
		\includegraphics[width=1.2\textwidth]{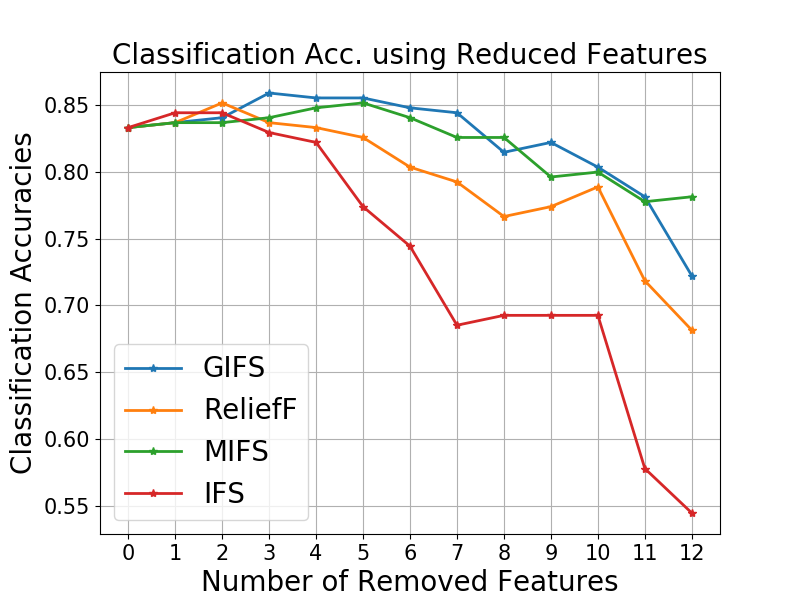}
		\parbox{1cm}{\small \hspace{3.5cm}(b)StatlogHeart}
	\end{minipage}
	\hspace{3ex}   %%两个minipage之间相隔3个字符的距离
	\begin{minipage}[t]{0.3\linewidth}
		\centering
		\includegraphics[width=1.2\textwidth]{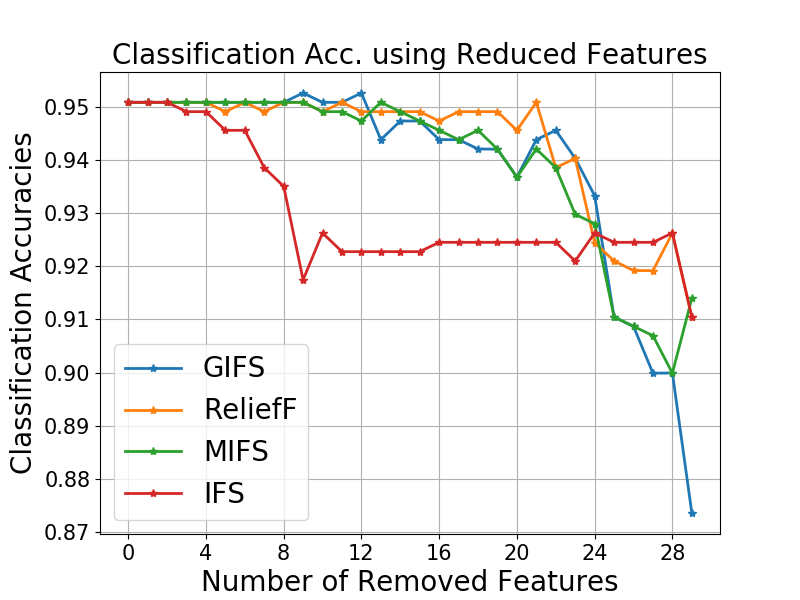}
		\parbox{1cm}{\small \hspace{3.5cm}(c)WDBC}
	\end{minipage}
	\caption{Performance comparison of different FS methods on three test datasets}
	\label{acc_reduce_features}
\end{figure*}

\section{Experiments \& Results}
The performance of the proposed method was evaluated for a binary classification problem using eight public datasets.

\subsection{Datasets}
\subsubsection{\textbf{Training Data Repository}}\label{training_repository}
Based on the description in Section \ref{configure_data}, 1000 datasets were generated by using randomly selected parameter values within the defined ranges in Table~\ref{madelon_parameters}. Meta features were then extracted from these data repository. Following the process described in Section~\ref{FS_Label} and \ref{metadata_construction}, the meta dataset for training was constructed, which contained 1000 data samples each had six meta features. The meta label of each sample was one of the four FS methods introduced in Section~\ref{FS_methods}. 

\subsubsection{\textbf{Testing Data Repository}}
Eight binary classification datasets which come from the UCI machine learning repository \cite{Dua:2019} were used to evaluate the performance of the proposed method. The detailed information is shown in Table~\ref{test_data}.
In Table~\ref{test_data}, \#Fea. and \#Samples represent the number of features and samples of the dataset, respectively. The distribution over class means the number of samples for each of the binary classes.

\begin{table}[tb]
	\caption{Description of the biomedical datasets for testing}
	\centering
	\begin{tabular}{c c c c c}
		\toprule
		\textbf{Dataset} & \textbf{\#Fea.} & \textbf{\#Samples} & \textbf{Distribution Over Class}  \\
		\midrule
		\centering
		Appendicitis  &  7 &  106  &  85 / 21     \\
		PIMA 		  &  8 &  768  &  500 / 268   \\
		WBC			  &  9 &  699  &  458 / 241   \\
		Statlog Heart & 13 &  270  &  150 / 120   \\
		Parkinsons	  & 22 &  195  &  48 / 147    \\
		WDBC          & 30 &  569  &  212 / 357   \\
		Spectfheart   & 44 &  267  &  55 / 212    \\
		Sonar         & 60 &  208  &  97 / 111    \\
		\bottomrule
	\end{tabular}
	\label{test_data}
\end{table}

\subsection{Feature Selection Methods}\label{FS_methods}
Four filter FS methods which come from different categories~\cite{li2017feature} were implemented in this experiment, i.e. Gini Index FS (GIFS)~\cite{singh2010feature}, ReliefF~\cite{robnik2003theoretical}, Mutual Information FS (MIFS)~\cite{battiti1994using} and Infinite FS (IFS)~\cite{roffo2015infinite}. Logistic regression was used to evaluate the algorithms' classification performance using the generated feature rankings by different FS methods. Through implementing different FS methods using the metric described in Section~\ref{FS_Label} on the training repository, the number of best performances achieved by each FS method was 546, 196, 147 and 111, respectively (total of 1000 datasets).

\subsection{Comparison of the Features' Distribution}
The distributions of the meta features from both 1000 training and eight testing datasets are shown in Fig.~\ref{compare_distribution}. It can be seen that the distributions of the meta features in the training repository cover the value range of the test datasets well for meta feature NS, NF and ACVF. Meta feature AAF, ACF and AEF of the test datasets are slightly higher or lower than the corresponding value ranges in the training datasets. This could be further improved by fine tuning the parameters in the data synthesis procedure.

\subsection{Evaluation Results}
We firstly applied individual FS methods (i.e. GIFS, ReliefF, MIFS and IFS) to the eight test datasets. The classification accuracies by gradually removing the least significant features for PIMA, Statlog Heart and WDBC datasets are shown in Fig.~\ref{acc_reduce_features}.

It can be seen that different FS methods had significantly different behaviours, which are difficult to be quantified and compared. Hence, we used our proposed measurement metric WS in (\ref{weighted_acc}) as the evaluation metric. The evaluation results by applying each of the four FS methods to the test datasets are listed in Table \ref{weighted_sum_acc}. The FS method that produced the highest WS value was treated as the ground truth (`Best Method' column). The recommended FS method using our proposed framework is listed in the last column of Table \ref{weighted_sum_acc}.

\begin{table}[tb]\scriptsize
    \centering
    \caption{Performance comparison on 8 test datasets}
    \begin{threeparttable}
        \begin{tabular}{|c|c|c|c|c|c|c|}
            \hline
            \textbf{Datasets}     & \textbf{GIFS}   & \textbf{ReliefF}     & \textbf{MIFS}  & \textbf{IFS}    & \textbf{\begin{tabular}[c]{@{}c@{}}Best\\ Method\end{tabular}} & \textbf{\begin{tabular}[c]{@{}c@{}}Recommend\\ Method\end{tabular}} \\ \hline
            \textbf{Appendicitis} & \textbf{2.41}  & 2.40   & \textbf{2.41} & 2.40  & MIFS  & MIFS  \\ \hline
            \textbf{PIMA}         & $2.63'$  & 2.62  & \textbf{2.64} & 2.56  & MIFS & MIFS          \\ \hline
            \textbf{WBC}          & $3.78'$  & 3.78  & \textbf{3.79} & 3.78  & MIFS & MIFS          \\ \hline
            \textbf{Statlog Heart}  & $4.84'$  & 4.61  & \textbf{4.85} & 4.08  & MIFS   & MIFS      \\ \hline
            \textbf{Parkinsons}   & \textbf{8.52}  & 8.29   & $8.49'$  & 8.42  & GIFS   & ReliefF   \\ \hline
            \textbf{WDBC}         & 13.48    & \textbf{13.60} & $13.51'$    & 13.41  & ReliefF  & ReliefF  \\ \hline
            \textbf{Spectfheart}  & 16.03    & $16.07'$     & 15.85  & \textbf{16.11} & IFS     & MIFS     \\ \hline
            \textbf{Sonar}        & \textbf{16.07} & $15.96'$   & 15.31 & 12.83  & GIFS  & ReliefF         \\ \hline
        \end{tabular}    
		\begin{tablenotes}
			\footnotesize
			\item Bold numbers indicate the best performance; 
			\item Numbers with $'$ indicate the second best performance.
		\end{tablenotes}    
    \end{threeparttable}
    \label{weighted_sum_acc}
\end{table}

It can be observed that our proposed method successfully recommended the best method for the Appendicitis, PIMA, WBC, Statlog Heart and WDBC dataset. In the case of the Sonar dataset, our recommended method ranked the second, which was only slightly lower than the best method. In the other two datasets (i.e. Parkinsons and Spectfheart), the proposed method cannot accurately recommend the best method. This may be due to the fact that the meta feature distributions of these datasets are outside the value ranges of the training dataset. This could be further improved by refining the data synthesis process in Section~\ref{configure_data}.

\begin{figure}[tb]
	\centering
	\includegraphics[trim=40 0 40 0, clip, width=\columnwidth]{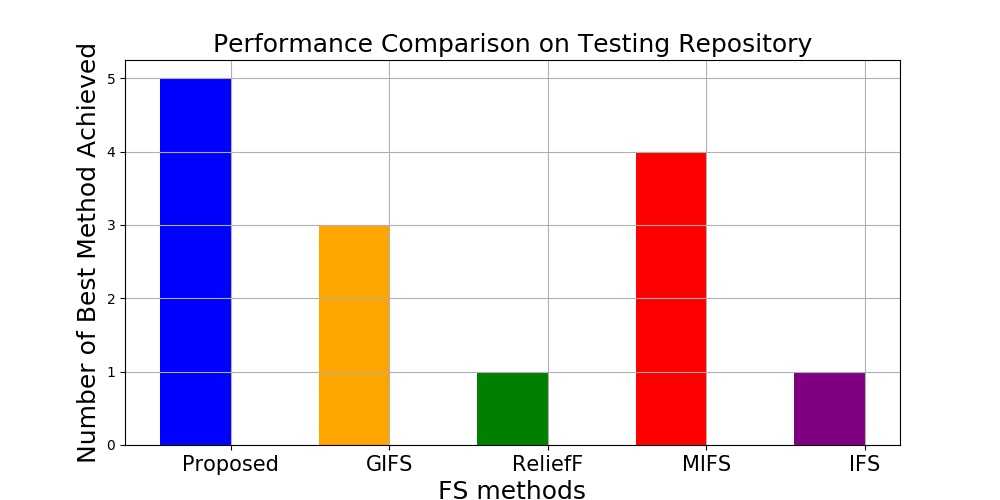}
	%\decoRule
	\caption[Framework of Fuzzy Sets Generation using Bootstrap Sampling]{Performance comparison on testing repository}
	\label{best_method_achieved}
\end{figure}

By counting the number of the achieved best method in the testing data repository, the performance between our proposed method and the individual FS methods are compared and displayed in Fig.~\ref{best_method_achieved}. 

It can be seen that our proposed method achieved the best performance comparing with the other individual FS methods. In the testing data repository, our proposed method successfully recommended the best method on five datasets out of the eight in total. In contrast, the individual FS methods only achieved the best performance in three, one, four and one cases, respectively. Overall, the successful recommendation rate of our proposed method was 62.5\% on the testing repository.

\subsection{Computational Cost}
In this section, we report and compare the execution time using different FS methods and the proposed method. The programs were implemented using Python and ran on a laptop with 2.2GHz, Intel(R) Core(TM) i5-5200U CPU and 8GB RAM. Each method was implemented and ran 10 times. The average execution time (s) of each method is reported in Table \ref{computational_cost}.

\begin{table}[tb]\scriptsize
    \caption{Average run time using different methods (/s)}
    \begin{tabular}{|c|c|c|c|c|c|c|}
    \hline
    \multirow{2}{*}{\textbf{Datasets}} & \multicolumn{4}{c|}{\textbf{Individual Methods}}  & \multirow{2}{*}{\textbf{\begin{tabular}[c]{@{}c@{}}Meta\\Learning\end{tabular}}} & \multirow{2}{*}{\textbf{\begin{tabular}[c]{@{}c@{}}Total Run\\Time\end{tabular}}} \\ \cline{2-5}
    & \textbf{GIFS} & \textbf{ReliefF} & \textbf{MI} & \textbf{IFS} &   &    \\ \hline
    \textbf{Appendicitis}  & 0.37  & 0.36  & 0.32  & 0.09  & 0.07  & 0.39  \\ \hline
    \textbf{PIMA}          & 1.55  & 7.50  & 0.99  & 0.28  & 0.32  & 1.31  \\ \hline
    \textbf{WBC}           & 2.71  & 10.19 & 2.61  & 2.68  & 1.30  & 3.91  \\ \hline
    \textbf{Statlog Heart} & 0.56  & 1.30  & 5.66  & 6.19  & 0.08  & 5.74  \\ \hline
    \textbf{Parkinsons}    & 2.55  & 1.04  & 1.21  & 0.43  & 0.38  & 1.42  \\ \hline
    \textbf{WDBC}          & 15.06 & 6.23  & 3.97  & 1.86  & 2.03  & 8.26  \\ \hline
    \textbf{Spectfheart}   & 3.40  & 3.16  & 4.18  & 2.23  & 0.20  & 4.37  \\ \hline
    \textbf{Sonar}         & 7.66  & 1.74  & 3.36  & 1.29  & 0.72  & 2.46  \\ \hline
    \end{tabular}
    \label{computational_cost}
\end{table}

The figures for the individual methods presented in Table \ref{computational_cost} show the average run time for the implementation of each FS method. The `Meta Learning' column indicates the execution time of our proposed framework. In addition, total run time represents the summation of the execution time using our meta learning framework and the recommended FS method accordingly.

It can be seen that our meta learning framework takes less than one second to run in most cases. By inspecting the total run time, the proposed method has not led to a significantly longer overall execution time, which indicates a high computational efficiency. Comparing with the individual FS methods, our meta learning framework and the recommended FS method has just consumed a moderate amount of time. This indicates that there is comparatively little additional computational cost incurred in implementing our meta learning framework, indicating the applicability of the approach. The use of our meta learning method provides an efficient way to learn the potentially optimal FS method.

\section{Discussion}
The results show that our method has successfully recommended the most appropriate FS method to use in five out of eight evaluation datasets. The overall successful recommendation rate was 62.5\% on the testing repository. From Table \ref{weighted_sum_acc}, it can be seen that the correct recommendations often appear on the datasets with small performance difference between various methods. This may be the misconception caused by the lack of the testing datasets with diverse performance. Further work need to be done to better evaluate it.

As for the computational cost, our proposed method is fast to run and so is sufficiently fast to be able to be widely used (apart from in situations which are very time critical). Rather than choosing one single FS method randomly, the pre-selection process using our meta learning framework has introduced very small additional computational burden. This actually makes it an attractive potential method to be used when a wide variety of candidate algorithms are considered.

Generally speaking, the absolute performance of our proposed method is not very remarkable. However, the point of this current paper is to demonstrate the potential of the method and the overall framework. The results are still provisional and clearly need to be improved in the future, if the approach is to be more widely used.

\section{Conclusion}
In this paper, we have proposed and implemented a meta learning method to recommend the best FS method from four candidate methods using a fuzzy similarity measure based framework. 

Instead of using real datasets for meta learning, we generate 1000 different synthetic datasets to form the training repository. Six meta features are extracted from the training data repository. The FS methods' performance is measured by using a novel weighted sum classification accuracy measurement. Based on the constructed meta datasets, a fuzzy similarity measure based framework is then applied to train a classification model. By evaluating the proposed method on eight different datasets from real-world applications, our proposed method successfully recommended the best method for five of these datasets, which is better than any of the individual FS method. Besides, our proposed method is computationally efficient with almost no additional time cost to the feature selection process.

For future work, we will extend our proposed method by generating better training data repository with wider distributions, introducing more meta features and testing other evaluation metrics for FS. More comparisons will be performed to better evaluate the proposed framework, such as using different machine learning methods, various FS methods, datasets with diverse performance, recent meta learning models and etc.

\clearpage

% The fuzzy similarity measure based framework will also be compared with other machine learning methods. \hl{}

% \section*{Acknowledgment}

% The preferred spelling of the word ``acknowledgment'' in America is without 
% an ``e'' after the ``g''. Avoid the stilted expression ``one of us (R. B. 
% G.) thanks $\ldots$''. Instead, try ``R. B. G. thanks$\ldots$''. Put sponsor 
% acknowledgments in the unnumbered footnote on the first page.

% \section*{References}

\bibliographystyle{IEEEtran}
\bibliography{sample}

\end{document}